\documentclass[runningheads]{llncs}
\usepackage{accv}
\usepackage{accvabbrv}
\usepackage{graphicx}
\usepackage{booktabs}

\usepackage[accsupp]{axessibility} 
\usepackage[pagebackref,breaklinks,colorlinks,citecolor=accvblue]{hyperref}
\usepackage{orcidlink}

\begin{document}

\title{Tracking Reflected Objects: A Benchmark}

\author{Xiaoyu Guo$^{\dagger}$ \and
Pengzhi Zhong$^{\dagger}$ \and
Lizhi Lin \and
Hao Zhang \and
Ling Huang \and
Shuiwang Li\thanks{Corresponding author. $^{\dagger}$These authors contributed equally.}
}
\authorrunning{Guo et al.}
\institute{College of Computer science an d Engineering,\\
Guilin University of Technology
, China \\
\email{\{guoxiaoyu, zhongpengzhi\}@glut.edu.cn, \{linlizhi0204, zhanghao\_0618, huangling240312, lishuiwang0721\}@163.com}
}

\maketitle

\begin{abstract} 

Visual tracking has advanced significantly in recent years, mainly due to the availability of large-scale training datasets. These datasets have enabled the development of numerous algorithms that can track objects with high accuracy and robustness.
However, the majority of current research has been directed towards tracking generic objects, with less emphasis on more specialized and challenging scenarios. One such challenging scenario involves tracking reflected objects. Reflections can significantly distort the appearance of objects, creating ambiguous visual cues that complicate the tracking process. This issue is particularly pertinent in applications such as autonomous driving, security, smart homes, and industrial production, where accurately tracking objects reflected in surfaces like mirrors or glass is crucial. To address this gap, we introduce TRO, a benchmark specifically for Tracking Reflected Objects. TRO includes 200 sequences with around 70,000 frames, each carefully annotated with bounding boxes. This dataset aims to encourage the development of new, accurate methods for tracking reflected objects, which present unique challenges not sufficiently covered by existing benchmarks. 
We evaluated 20 state-of-the-art trackers and found that they struggle with the complexities of reflections. To provide a stronger baseline, we propose a new tracker, HiP-HaTrack, which uses hierarchical features to improve performance, significantly outperforming existing algorithms. We believe our benchmark, evaluation, and HiP-HaTrack will inspire further research and applications in tracking reflected  objects. The TRO and code are available
at \url{https://github.com/OpenCodeGithub/HIP-HaTrack}.

  \keywords{Reflected object tracking \and Benchmark \and Hierarchical feature aggregation}
\end{abstract}

\section{Introduction}
\label{sec:intro}

The field of visual tracking has experienced significant advancements over the past decade, largely due to the increasing availability of large-scale datasets and the development of sophisticated algorithms \cite {3,4,5,8,69,71,72,73}. These advancements have enabled tracking systems to achieve remarkable accuracy and robustness in a variety of applications, such as autonomous driving, surveillance, and smart home technologies \cite{2,69}. With improved algorithms and richer datasets, visual tracking systems can now better handle occlusions, variations in object appearance, and environmental changes, which are common challenges in real-world scenarios \cite{3,4,5,6,7,8,9}. Despite these achievements, much of the research in visual tracking has been concentrated on generic objects, leaving more specialized and challenging scenarios relatively underexplored \cite{68,Wu2024TrackingTO}. 
One particularly challenging scenario that has not received adequate attention is the tracking of reflected objects. Reflections can severely distort the appearance of objects, creating ambiguous visual cues that complicate the tracking process. The reflective properties of surfaces like mirrors and glass can lead to multiple, overlapping images of the same object, making it difficult for algorithms to discern the true object from its reflection \cite{Ittelson1991ThePO,turnbull1996failure,bianchi2008relationship,gregory2010mirror,Sareen2015ThroughTL,Nightingale2019CanPD}.
Accurately tracking objects in reflections—whether in mirrors, glass surfaces, or other reflective materials—is essential for improving monitoring systems in various applications like surveillance, smart home, medical procedures, and traffic. 

\begin{figure*}[h]
	\centering
	{
		\begin{minipage}[t]{1\textwidth}
			\centering
			\includegraphics[width=1\textwidth,height=0.15\textwidth]{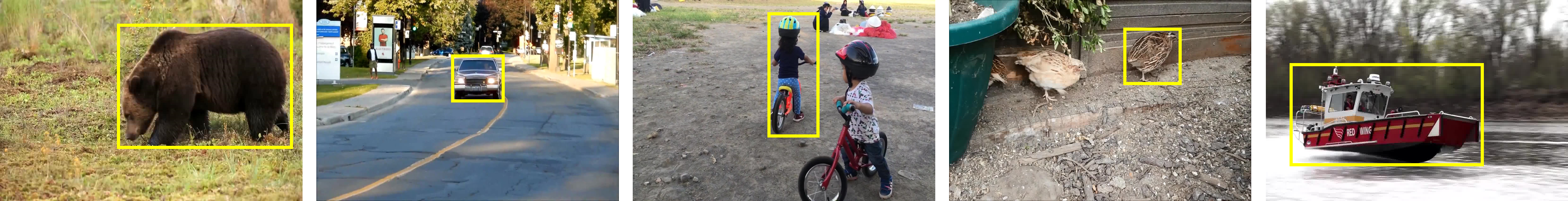}
			\centerline{\footnotesize (a) Example of generic object tracking.}
	\end{minipage}}
	{
		\begin{minipage}[t]{1\textwidth}
			\centering
			\includegraphics[width=1\textwidth,height=0.15\textwidth]{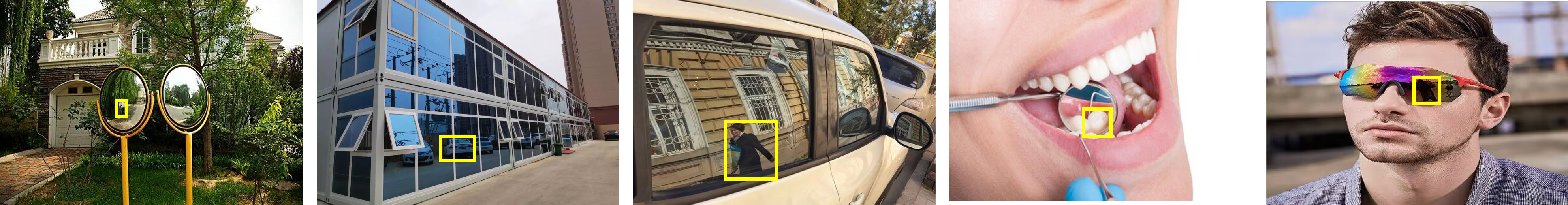}
			\centerline{\footnotesize (b) Example of reflected objects tracking.}
	\end{minipage}}
	\caption{Generic object tracking (a) and reflected objects tracking (b). Compared with generic object tracking, tracking of reflected objects is more challenging as their tendency to distort the appearance of objects and create ambiguous visual cues. }
	\label{fig:TRO_examples}\vspace{-0.25in}
\end{figure*}

In surveillance, surveillance systems often face challenges due to obstructions or blind spots that prevent full coverage. By leveraging mirror reflections, "virtual" cameras can be introduced to capture images through reflections, filling these blind spots and providing more comprehensive monitoring \cite{Shen2018BlindSM,Singhal2022ECMSEC,Ray2022ThePB}. In smart homes, reflective surfaces such as mirrors, windows, and polished furniture are ubiquitous. Correctly tracking objects reflected in these surfaces can significantly enhance the functionality and reliability of automated systems \cite{hossain2007smart,freysteinson2010assessing,bianchi2015differences,Latif2020SmartMF,kilic2024user}. In smart home, home automation systems with accurate tracking can enhance user experience and safety. By distinguishing between reflections of humans and objects, these systems can make better decisions about turning on lights, adjusting temperature, or locking doors, ensuring they respond to actual human presence \cite{saad2014smart,sovacool2020smart,chakraborty2023smart}. Smart home assistants can also use reflections to improve awareness, such as detecting a person entering a room through a mirror's reflection and offering a personalized greeting or adjusting the environment to the user's preferences \cite{stojkoska2017review,marikyan2019systematic,sovacool2020smart,chakraborty2023smart}. In the medical field, mirror reflections can provide crucial assistance in tracking surgical instruments and monitoring patients' vital signs \cite{bichlmeier2009virtual,poh2011medical,soppimath2019smart,jasmine2020medical}. During surgeries, reflections can offer additional angles and perspectives, allowing medical staff to have a comprehensive view of the operating area. This enhanced visibility can improve precision and safety, ensuring that instruments are used correctly and efficiently \cite{bichlmeier2006tangible,bichlmeier2009virtual,roy2020situs}. Similarly, monitoring vital signs through reflections can offer real-time feedback, helping medical professionals respond promptly to any changes in a patient's condition. These extra perspectives are vital for the accuracy and effectiveness of medical procedures, contributing to better patient outcomes and enhanced surgical performance \cite{bichlmeier2006tangible,jansen2018should,abi2019nonbiological}. In traffic monitoring, strategically placed mirrors at streetlights or intersections can significantly enhance surveillance capabilities \cite{mole2017looking,baysal2023assessment}. These mirrors capture reflected images of vehicles and pedestrians, effectively expanding the range of monitored areas. By using reflections, traffic monitoring systems can gain visibility into spots that are typically out of the direct line of sight of cameras, such as blind corners or areas obscured by other objects \cite{dhalwar2019image,sadrhaghighi2022monitoring}. This expanded field of view allows for more comprehensive coverage, ensuring that all traffic movements are accurately tracked. The improved detection accuracy helps in better traffic management, accident prevention, and enforcement of traffic laws \cite{diaz2003optical,mittal2017comparative,zhang2022traffic,li2023traffic}. Enhanced monitoring through reflections can also aid in incident response, providing valuable information about traffic conditions and potential hazards in real time \cite{loizides2018efficient,noriaki2023collision}.

Given these diverse applications and the critical nature of accurate tracking in these scenarios, it is clear that the challenges posed by reflected objects require dedicated research efforts. Current tracking algorithms, designed primarily for generic objects, often lack the robustness needed to handle these complexities effectively. In light of these challenges, there is a pressing need for the development of specialized benchmarks and algorithms tailored to the tracking of reflected objects. 
In this work, we introduce the Tracking Reflected Objects (TRO) benchmark, a dedicated dataset designed to stimulate research in this challenging area. TRO comprises 200 sequences, totaling approximately 70,000 frames, each meticulously annotated with bounding boxes. Each sequence undergoes semi-automatic object annotation, possessing various attributes for performance evaluation and analysis. This comprehensive dataset captures a wide range of real-world scenarios involving reflections, providing a robust foundation for developing and evaluating new tracking algorithms tailored to these conditions.
In addition, we present a detailed evaluation of 20 state-of-the-art tracking algorithms using the TRO benchmark. Our findings reveal that current methods often struggle with the complexities introduced by reflections, indicating a significant need for improvement. By releasing TRO, we aim to foster advancements in tracking technology that address the unique challenges of reflected objects, ultimately enhancing the accuracy and reliability of visual tracking systems across various applications. The main contributions of this paper are as follows:
\begin{itemize} 

\item We introduce TRO, which is the first benchmark dedicated to track reflected objects to our knowledge.  TRO, containing 200 sequences with meticulous annotations, poses unique challenges not adequately covered by existing benchmarks and may inspire further research and applications in tracking reflected objects.

\item We conduct a thorough evaluation of 20 state-of-the-art trackers on TRO. This evaluation contributes to our understanding of the limitations of current trackers, establishes benchmarks for performance comparison and inspires future research endeavors in the field of tracking reflected objects.

\item We propose HiP-HaTrack, a novel tracker that aggregate hierarchical features to enhance performance. HiP-HaTrack surpasses existing state-of-the-art algorithms in TRO, providing a stronger baseline for future research in tracking reflected objects.
\end{itemize}

\section{Related Works}
\subsection{Visual Tracking Algorithms}

There are generally two categories of mainstream visual trackers: DCF-based and DL-based trackers \cite{lu2019online,mazzeo2019visual}. DCF-based trackers treat visual tracking as an online regression problem. Thanks to the utilization of the Parseval theorem and Fast Fourier Transform (FFT), DCF-based trackers demonstrate high speed on CPUs \cite{li2021learning}. They initially began with the MOSSE (Minimum Output Sum of Squared Error) filter \cite{bolme2010visual}, and since then, significant progress has been made. For example, \cite{li2015scale,danelljan2016discriminative} employed an additional scale filter to handle target scale variations. \cite{danelljan2015learning,li2018learning} explored regularization techniques to enhance robustness, and \cite{li2020asymmetric} extended the DCF framework to achieve translation equivariance. However, the limited expressive power of manually crafted features makes it difficult for DCF-based trackers to maintain robustness under challenging conditions. As a result, DL-based trackers have gained significant attention for their remarkable capability to automatically learn features through neural networks. SiamFC \cite{bertinetto2016fully}, a typical representative of DL-based trackers, pioneered the utilization of the Siamese networks \cite{chicco2021siamese} in visual tracking and has spurred the development of various sophisticated trackers. For instance, those mentioned in \cite{li2019siamrpn++,guo2020siamcar,xu2020siamfc++,46,48,43} have demonstrated significant advancements in tracking precision and robustness.

Recently, Vision Transformers (ViTs) \cite{dosovitskiy2021an} have shown significant potential in streamlining and unifying frameworks for visual tracking, as evident in studies like\cite{dosovitskiy2021an,liu2021Swin,ye2022ostrack,36}. Based on ViTs, OStrack \cite{ye2022ostrack} combines feature extraction and relation modeling, introducing an early elimination module to implement a one-stream tracking framework. DropTrack \cite{36} improves temporal correspondence learning by adaptively performing spatial attention dropout during frame reconstruction. Very recently, HIPTrack \cite{33} offers a historical prompt network constructed with refined historical foreground masks and target visual features to provide comprehensive and precise guidance for the tracker. However, most current research focuses on tracking generic objects, with less attention given to more specialized and challenging scenarios, particularly the tracking of reflected objects. One of the main reasons for this is the lack of dedicated benchmark tests specifically designed for tracking reflective objects.

\subsection{Visual Tracking Benchmarks}
The development of visual tracking technology relies on accurate measurement and evaluation of tracking method performance, with benchmark datasets serving as crucial standards for evaluation. Currently, there are two main types of benchmark datasets: general benchmarks and specific benchmarks \cite{55}.

\textbf{General benchmarks.} General benchmark datasets are primarily used to evaluate tracking algorithm performance in general scenarios. OTB-2013 \cite{20}, a representative dataset, initially contained 50 sequences, later expanded to OTB-2015 \cite{21}, covering 100 sequences. VOT \cite{57} organization hosts a series of tracking competitions with up to 60 sequences. NFS \cite{58} focuses on high frame rate videos. Many large-scale benchmarks have been proposed to train DL-based trackers. TrackingNet \cite{61} aimed at providing resources for training and evaluating general object tracking algorithms,featuring over 30,000 sequences. GOT-10k \cite{62} contains over 10,000 video sequences, with each sequence containing rich tracking challenges. LaSOT \cite{12} initially contained 1,400 long-term video sequences and later added 150 additional sequences in subsequent studies, introducing a new evaluation scheme focusing on handling situations where the object becomes invisible during tracking.

\textbf{Specific benchmarks.} Specific benchmark datasets are designed to evaluate the tracking of particular objects. For example, the UAV123 \cite{mueller2016benchmark} focusing on low-altitude drone object tracking with 123 sequences captured by drone. VOT-TIR \cite{64} utilizes both RGB and thermal infrared images simultaneously to improve object tracking performance in RGB-T sequences. CDTB \cite{65} and PTB \cite{66} aim to evaluate tracking performance in RGB-D videos, where D represents depth images. TOTB \cite{55} focuses on transparent object tracking, collecting 225 videos from 15 transparent object categories. Recently, some new benchmarks have emerged, such as TSFMO \cite{68}, specifically objecting tracking small objects and fast-moving objects.  

\begin{figure*}[h]
	\centering
	\includegraphics[width=0.7\textwidth]{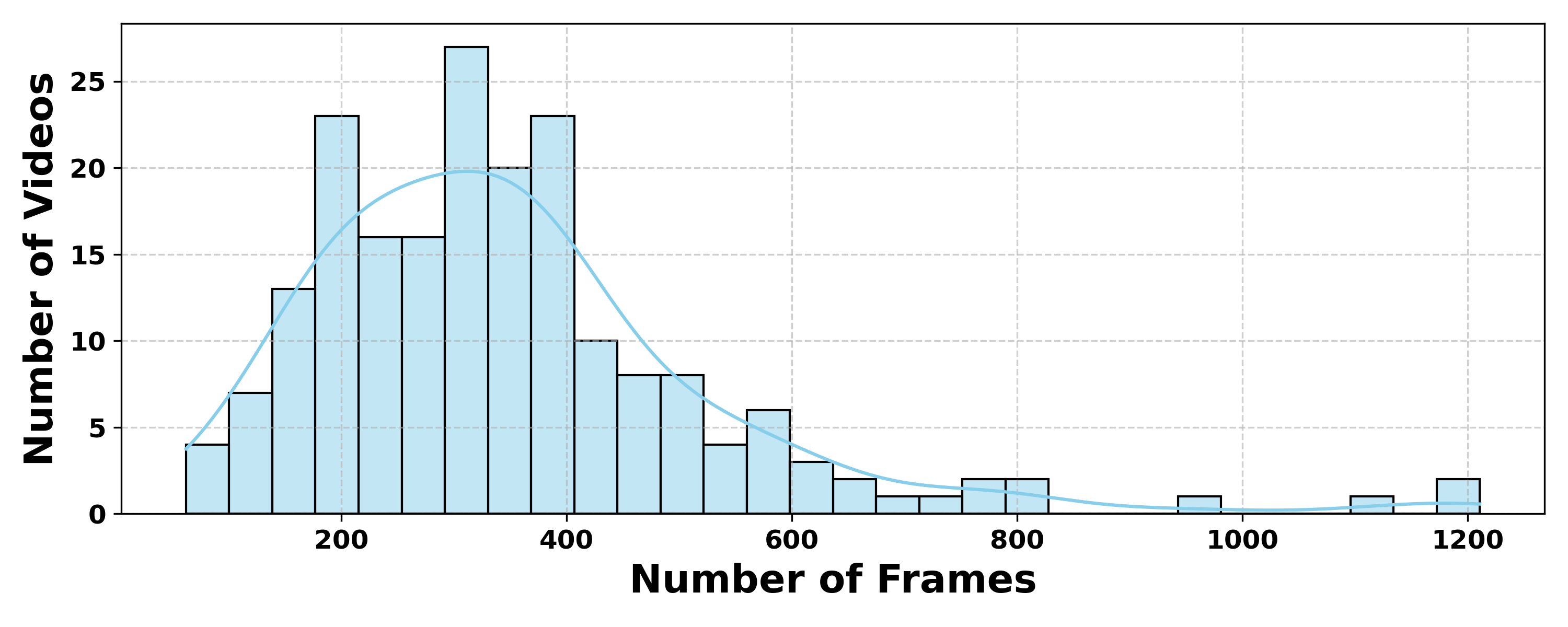}	
\caption{
This figure presents a histogram of the video frame counts in the TRO. The x-axis represents the number of frames per video, divided into bins, and the y-axis shows the number of videos within each bin. }
	\label{fig:2}	
\end{figure*}
However, existing benchmark datasets primarily focus on tracking physical objects. With the increasing use of reflected objects in fields such as industrial automation, security monitoring, and autonomous driving, the need for specialized tracking algorithms is growing. Reflected objects present unique challenges, including reflection, distortion, and ambiguity, which often cause traditional object tracking algorithms to perform poorly. Additionally, there is a notable lack of dedicated datasets for tracking reflected objects. This paper aims to address this by proposing a novel Reflected Object Tracking Dataset, designed to facilitate the development and evaluation of algorithms specifically for reflected object tracking.
\begin{table}[h]
\caption{Summary of statistics of the proposed TRO.}
\label{TRO_statistic}
\centering
\begin{tabular}{ll|ll|ll} 
			\toprule
			Number of videos &200 & Min frames &62 & Max frames  &1211    \\ 
Total frames  & 69810 & Avg frames &349 & Frame range &1149  \\ 
			\bottomrule
		\end{tabular}
\end{table}
\section{Construction of the TRO Dataset}

\subsection{Data Collection}

We collected hundreds of candidate videos through the internet and by recording with a camera. After that, we ranked and filtered the videos based on four criteria of tracking difficulty and some additional challenging cases. Vide os can gain higher rankings through the following ways: 1) sufficient relative motion of the target, 2) high variability of the environment, 3) the target crossing frame borders, and 4) sufficient footage length. In addition to the above criteria, videos with additional challenges are given higher priority. For example, distinguishing targets from other highly similar objects is a challenge in object detection and tracking.
After filtering, videos are further selected and sampled
into sequences with a frame number threshold ($\leq1300$). Relatively stationary frames were manually discarded. 
Due to resource constraints in collecting, processing, and annotating video data, we chose 200 sequences for the TRO benchmark. This ensures each video is high quality and thoroughly annotated, while staying within our available time, budget, and manpower.
The targets mainly include humans (such as riders, pedestrians and dancers), animals (such as dogs,  cats and tigers), and rigid objects (such as cars, airplanes and boats). Our benchmark covers a wide range of targets with high diversity. Table \ref{TRO_statistic} summarizes the TRO dataset, and Fig. \ref{fig:2} presents a histogram of the video frame counts in the TRO. Additionally, to highlight the challenges associated with tracking reflected objects, we evaluate seven state-of-the-art trackers on TRO and two popular generic object tracking benchmarks for comparison. The AUC are presented in Table \ref{tab:3} Remarkably, these tackers' AUC on TrackingNet is more than 15.0\% higher than TRO, highlighting the significant challenges posed by tracking reflected object.

\begin{table*}[]
\centering

\caption{ AUC (\%) Comparison of state-of-the-art trackers on the TRO and generic object tracking benchmarks. }
\label{tab:3}
\vspace{-0.1in}
	\resizebox{4.2in}{0.3in}{
\begin{tabular}{@{}cccccccccc@{}}
\toprule
Dataset     & HIPTrack      & ROMTrack      & DropTrack     & ARTrack       & SeqTrack      & GRM           & SimTrack              \\\hline
TRO         & \textbf{68.9} & \textbf{67.5} & \textbf{67.1} & \textbf{62.1} & \textbf{66.6} & \textbf{63.5} & \textbf{61.5}      \\
TrackingNet \cite{13} & 84.5          & 84.1          & 84.1          & 85.1          & 83.9          & 84.0            & 83.4                      \\
LaSOT \cite{12}       & 72.7          & 71.4          & 71.8          & 72.6          & 71.5          & 69.9          & 70.5             \\\bottomrule           
\end{tabular}
}\vspace{-0.12in}
\end{table*}

\subsection{Data Annotation}
\begin{figure*}[t]
	\centering
	\includegraphics[width=1\textwidth]{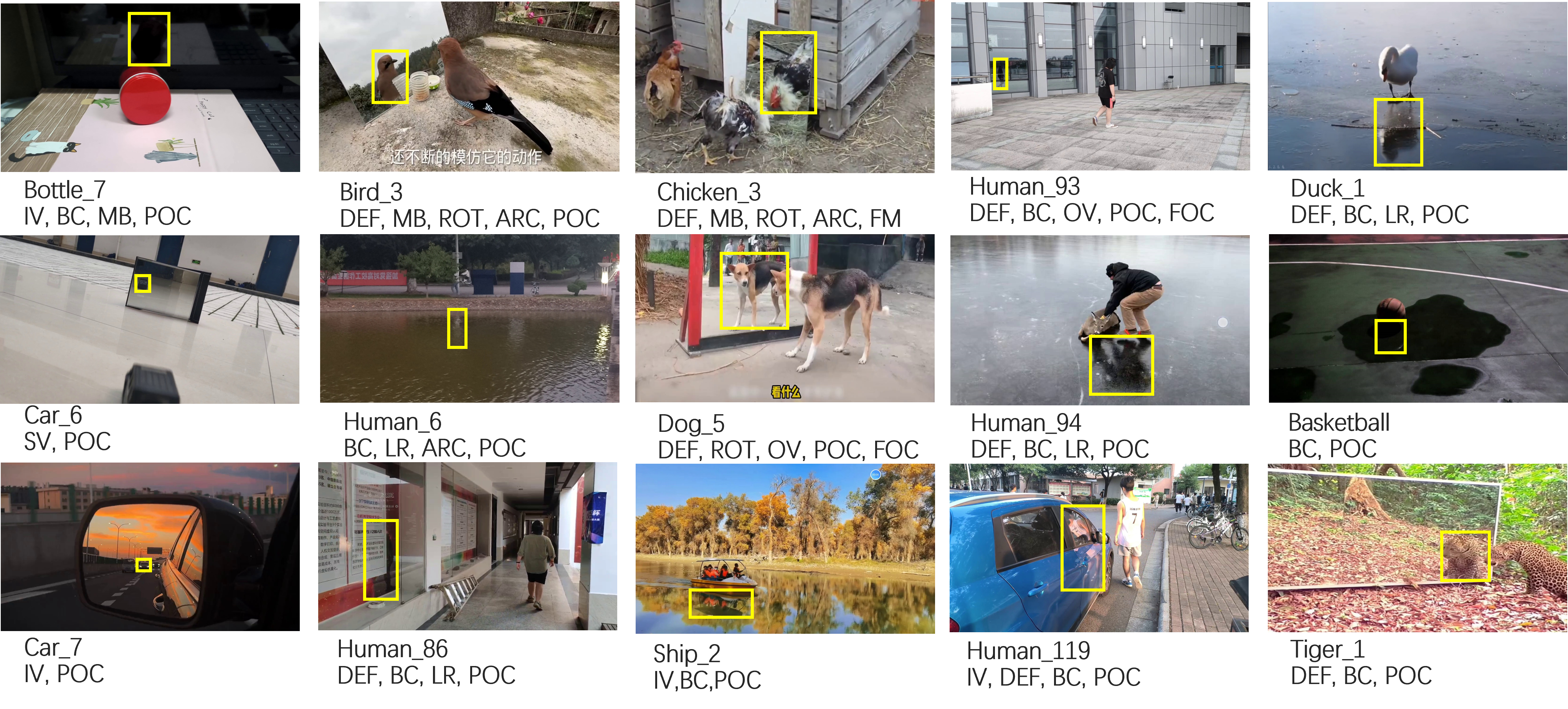}	\vspace{-0.25in}
	\caption{Some examples of box annotations for TRO.Each sequence is annotated with axis-aligned bounding boxes and attributes.}
	\label{fig:3}
\end{figure*}
We followed the principles outlined in \cite{14} for sequence annotation. Each video frame was meticulously labeled by a team of expert annotators (i.e., students working on visual tracking). Annotators drew/edited axis-aligned bounding boxes around the visible portion of the object in each frame, representing the tightest bounding box. Objects were labeled if visible, otherwise marked as absent, out of view, or occluded. To improve annotation efficiency and quality, we employed an efficient method called point annotation proposed in \cite{kirillov2023segment}. Objects were annotated in each frame using Segment Anything \cite{kirillov2023segment}, a versatile segmentation method that automatically identifies objects or regions in images and generates annotation information. And than, we conducted visual inspection of the automatically generated annotations from Segment Anything. For cases of poor quality, we performed manual annotation by reviewing and correcting errors or omissions. In addition, a double-check mechanism was implemented where a second annotator reviewed the initial annotations to maintain accuracy and consistency. Any discrepancies were resolved through discussion and consensus. This combination of automatic and manual annotation along with the double-check mechanism resulted in a high-quality annotated dataset, which serves as a reliable foundation for subsequent research. Our annotation methods and strategies ensure both efficiency and accuracy. Examples of box annotations in TRO can be seen in Fig. \ref{fig:3}.

\subsection{Attributes}
\begin{figure*}[t]
	\centering
	{
		\begin{minipage}[t]{0.48\textwidth}
			\centering
			\includegraphics[width=1\textwidth,height=0.58\textwidth]{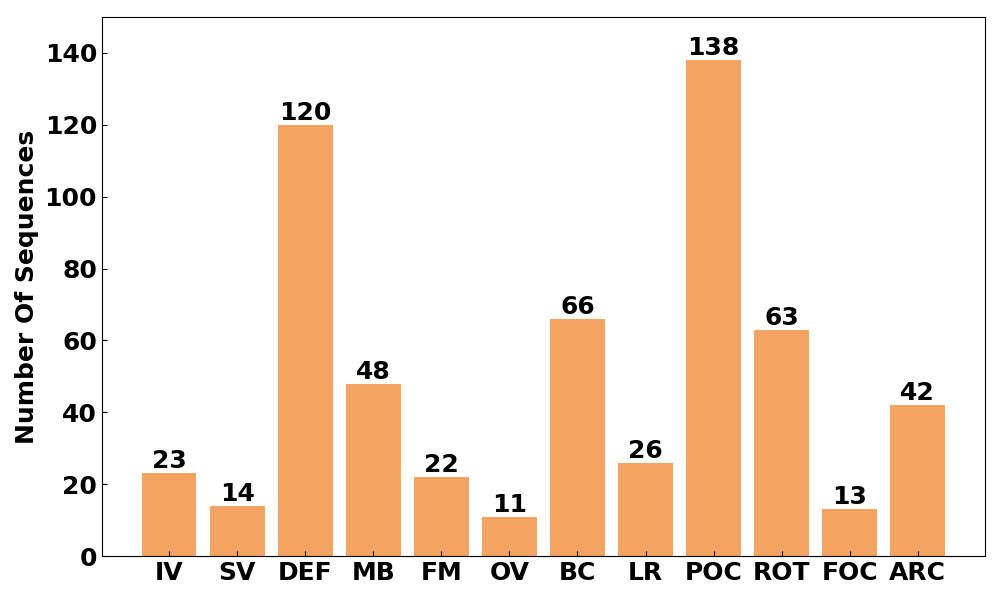}\hspace{0in}
   \centerline{\footnotesize (a)}			
	\end{minipage}}
	{
		\begin{minipage}[t]{0.48\textwidth}
			\centering
			\includegraphics[width=1\textwidth,height=0.58\textwidth]{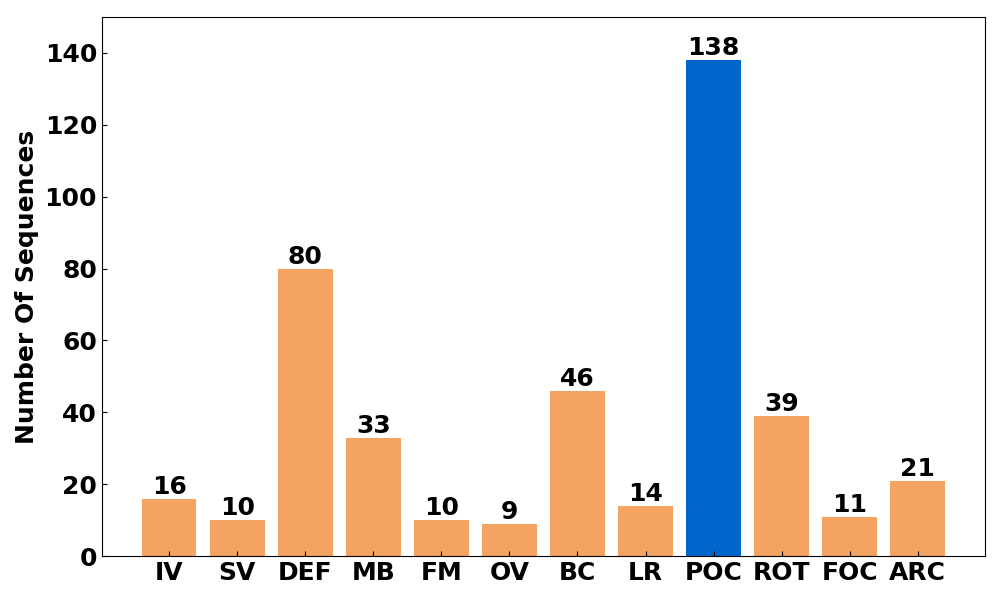}\hspace{0in}
   \centerline{\footnotesize (b)}
	\end{minipage}}
 \vspace{-0.15in}
	\caption{ (a) Attribute distribution of the entire testset, and (b) the
 distribution of the sequences with Partial Occlusion (POC) attribute. }
	\label{fig:attributes}\vspace{-0.15in}
\end{figure*}
In line with the approach of popular tracking benchmarks \cite{zhu2021detection},\cite{62}, and \cite{21}, we further annotated each video sequence with multiple attributes to better analyze the performance of tracking algorithms. The proposed TRO dataset features 12 attributes, including (1) Illumination Variation (IV), (2) Scale Variation (SV), (3) Deformation (DEF), (4) Motion Blur (MB), (5) Fast Motion (FM), which is assigned when the target center moves by at least 20\% of its size
in last frame. (6) Out-of-View (OV), (7) Background Clutter (BC), (8) Low Resolution (LR), assigned when the object region is less than 900 pixels, (9) Partial Occlusion (POC), (10) Rotation (ROT), (11) Full Occlusion (FOC), and (12) Aspect Ratio Change (ARC), which assigned when the aspect ratio of the bounding box exceeds the range [0.5, 2]. By considering these attributes, we can gain a more comprehensive understanding of the performance of tracking algorithms under different challenging scenarios.

The distribution of attributes in our dataset is shown in Fig. \ref{fig:attributes}(a). Some attributes occur more frequently, such as DEF (Deformation) and POC (Partial Occlusion). It also indicates that a sequence typically has multiple attribute annotations. In addition to summarizing the performance of the entire testset, we can evaluate tracking methods on a subset corresponding to a particular attribute to report performance on specific challenging conditions. For example, the POC subset contains 138 sequences, which can be used to analyze the performance of trackers in handling occlusions. 

The attribute distribution in the POC subset is shown in Fig. \ref{fig:attributes}(b), which indicates that the challenge associated with POC frequently co-occurs with other challenges rather than presenting alone. This overlapping of attributes highlights the complex nature of real-world tracking scenarios and underscores the need for robust algorithms capable of handling multiple simultaneous challenges. Such insights are crucial for developing and evaluating advanced tracking methods that can perform effectively in diverse and complicated conditions.

\section{A New Baseline: HIP-HaTrack}

During our evaluation, we observed that although HIPTrack outperforms other state-of-the-art trackers on TRO, but its performance remains significantly below satisfactory levels. Therefore, based on HIPTrack, we propose a novel baseline tracker called HIP-HaTrack based on aggregating hierarchical features to enhance the ability to address the challenges posed by significant ambiguous visual cues and distortion of object appearance caused by reflections. Reflections and ambiguous visual cues can cause certain features to become less reliable. 
As multi-level features capture information at various scales and resolutions, aggregating these features enables the tracker to rely on a broader set of data, thereby increasing the likelihood of correctly identifying the target. In addition, aggregating hierarchical features allows the tracker to adaptively utilize the most relevant features for a given scenario. 

\begin{figure*}[t]
	\centering
	\includegraphics[width=1.0\textwidth]{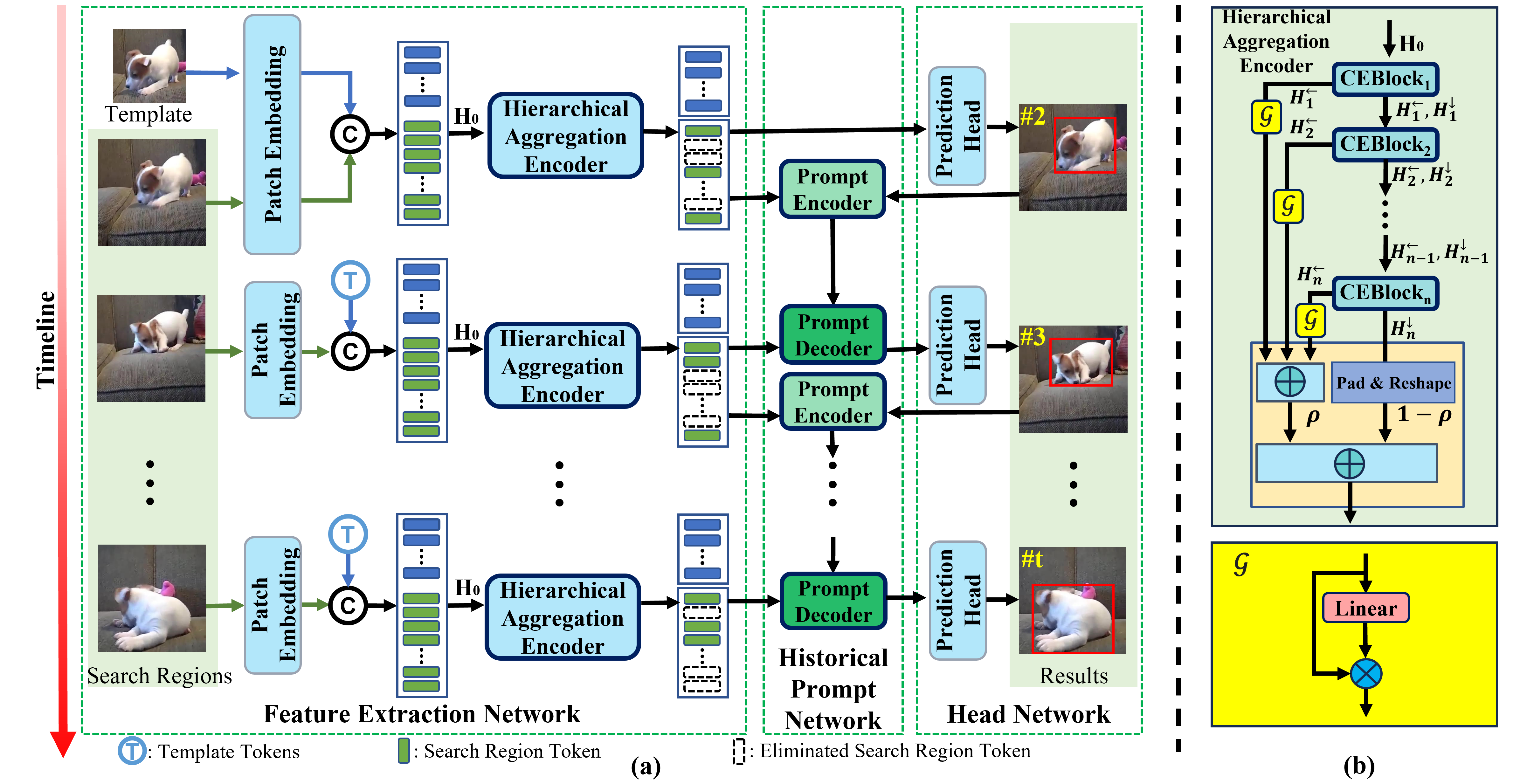}	\vspace{-0.15in}
	\caption{(a) The overall framework of HIP-HaTrack,which  inherited from HIPTrack. The difference lies in the encoder part of the feature extraction network. (b) The structure of the Hierarchical Aggregation Encoder.
}
	\label{fig:hip_net}	
\end{figure*}

\subsection{Overall Architecture}
Like HIPTrack, HIP-HaTrack also comprises three main components, as shown in Fig. \ref{fig:hip_net}(a). The feature extraction network is responsible for extracting features from the search region that contain target-template matching information, while filtering out background image patches. The historical prompt network contains a historical prompt encoder and a historical prompt decoder. The historical prompt encoder needs to encode the position information and visual features of the target in the current frame as historical target features, and store them in the memory bank of the historical prompt decoder. And the historical prompt decoder generates historical prompt for each search region and combines these historical prompt with the compressed features of the search region. The head network has a same structure to OSTrack \cite{ye2022ostrack}.The difference between HIP-HaTrack and HIPTrack lies in their feature extraction network. HIPTrack utilizes a Vision Transformer (ViT) with early candidate elimination modules, similar to the one used in OSTrack, serving as its feature encoder. Our HIP-HaTrack introduces a novel Hierarchical Aggregation Encoder, as illustrated in Fig. \ref{fig:hip_net}(b). This Hierarchical Aggregation Encoder is designed to integrate features at multiple levels of abstraction. By aggregating hierarchical features, HIP-HaTrack aims to enhance the feature representations, particularly for tracking reflected objects. In the following section, we will provide a detailed description of the proposed Hierarchical Aggregation Encoder. Other architectures inherit from HIPTrack; please refer to \cite{33} for details.

\subsection{Hierarchical Aggregation Encoder}
Our Hierarchical Aggregation Encoder builds upon the Transformer Encoder architecture of HIPTrack, namely a Vision Transformer (ViT) with early candidate elimination modules. This structure is represented with stacked CEBlocks as shown in Fig. \ref{fig:hip_net}(b). 
The key difference is the proposed aggregation structure for aggregating hierarchical features output by the ViT blocks. Each hierarchical feature is adaptively weighted with a gate module represented by $\mathcal{G}$. The weighted hierarchical features are summed to obtain an aggregated feature, which is finally weighted summed with the feature produced by the original Transformer Encoder. Considering that the early candidate elimination module in each CEBlock may alter the shape of the input feature, we divide the CEBlock into two parts: one representing the ViT block with early candidate elimination, denoted by $CBlock^{\downarrow}i$, and the other representing the ViT block without it, denoted by $CBlock^{\gets}i$. The output features produced by these two parts are represented by $H^{\downarrow}{i}$ and $H^{\gets}{i}$, respectively. The effect of the proposed Hierarchical Aggregation Encoder can be formally formulated by
\begin{equation}
F_{ha}=(1-\rho )\mathcal{P}(H^{\downarrow }_n)+\rho\sum_{i=1}^{n}\mathcal{G}(H^{\gets}_{i})\cdot H^{\gets}_{i},
\end{equation}
where $H^{\gets}_{i}=CBlock^{\gets}_i(H^{\gets}_{i-1})$, $H^{\downarrow }_{i}=CBlock^{\downarrow }_i(H^{\downarrow }_{i-1})$, $\mathcal{P}$ denotes the padding and reshaping used to recover the shape of $H^{\downarrow }_n$ to that of $H_0$, $\rho\in [0,1]$ is a weighting constant used to balance the importance of aggregated feature and $H^{\downarrow }_n$. Note that if $\rho=0$ $F_ha$ is just the feature produced by the Transformer Encoder of HIPTrack. Thus, our Hierarchical Aggregation Encoder extends and generalizes the feature extraction network of HIPTrack. It enhances feature representation by capturing information at multiple levels of abstraction, thereby improving the richness and comprehensiveness of the features. This approach also provides flexibility for further fine-tuning or extension, making it more effective for tracking objects affected by reflections. Since the proposed Hierarchical Aggregation Encoder does not introduce any additional training loss, we can utilize the same training pipeline as HIPTrack for training HIP-HaTrack. Please refer to HIPTrack \cite{33} for details.
\begin{figure*}[t]
	\centering
	{
		\begin{minipage}[t]{0.48\textwidth}
			\centering
			\includegraphics[width=1\textwidth,height=0.58\textwidth]{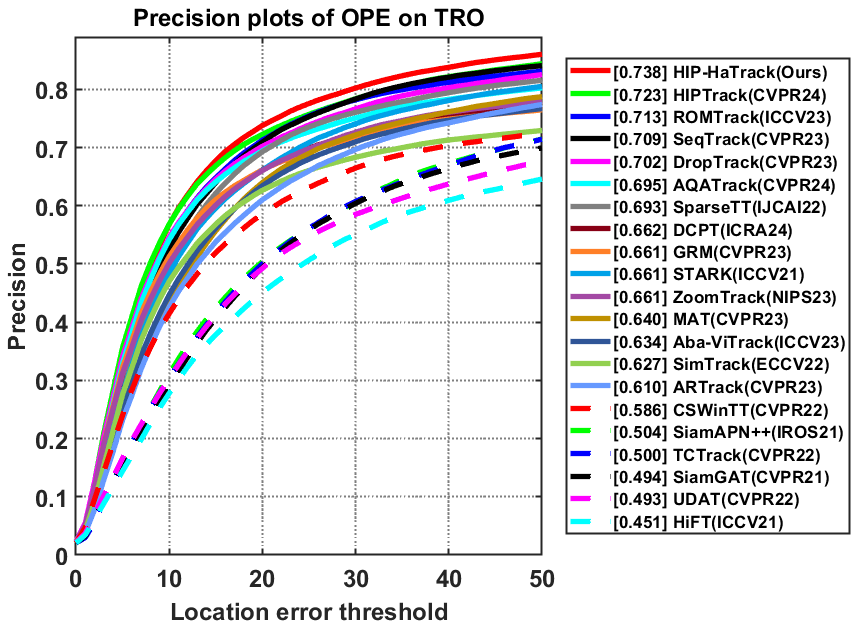}\hspace{0in}
			
	\end{minipage}}
	{
		\begin{minipage}[t]{0.48\textwidth}
			\centering
			\includegraphics[width=1\textwidth,height=0.58\textwidth]{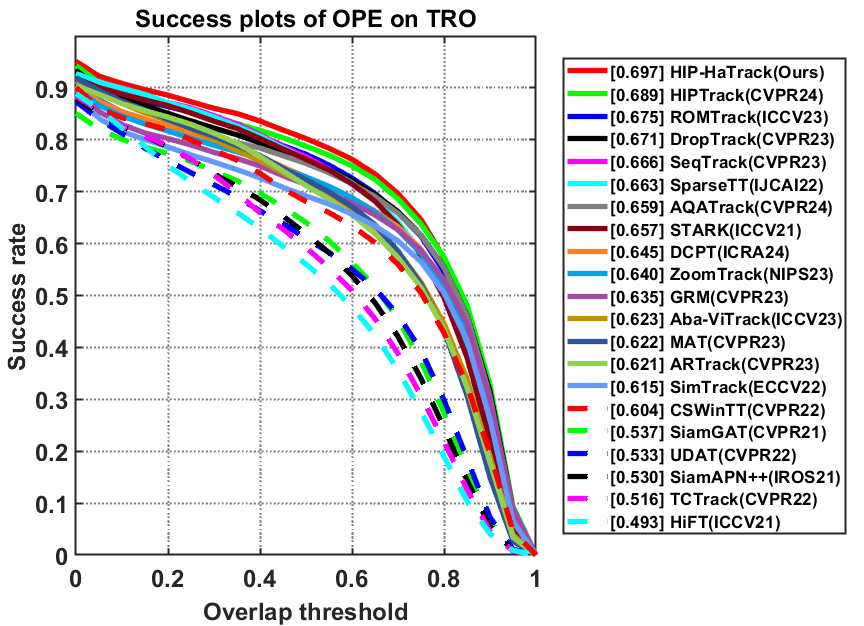}\hspace{0in}

	\end{minipage}}
 \vspace{-0.15in}
	\caption{Overall performance on TRO. Precision and success rate for one-pass evaluation (OPE) \cite{20} are used for evaluation. }
	\label{fig:TRo_OPE}\vspace{-0.15in}
\end{figure*}

\section{Evaluation}
In this work, precision and success rate are used as the quantitative metrics for evaluation.

\noindent\textbf{Precision Plot:} 
The precision plot is a common evaluation metric used in tracking studies. It typically assesses tracking precision through center position error, which quantifies the average Euclidean distance between the tracked object's center and its manually labeled ground truth position. This metric summarizes tracking performance by averaging errors across all frames in a sequence. However, when a tracker loses track of the object, its output position may become random, making the average error less reliable \cite{53}. To address this, the precision plot has gained popularity. It evaluates overall tracking performance by comparing the tracker's estimated position with the ground truth within a specified distance threshold \cite{53,54}. For our evaluation, we adopt the precision score using a 20-pixel threshold, following established practices \cite{53}.
\begin{figure*}[h]
	\centering
	{
		\begin{minipage}[t]{0.48\textwidth}
			\centering
			\includegraphics[width=1\textwidth,height=0.58\textwidth]{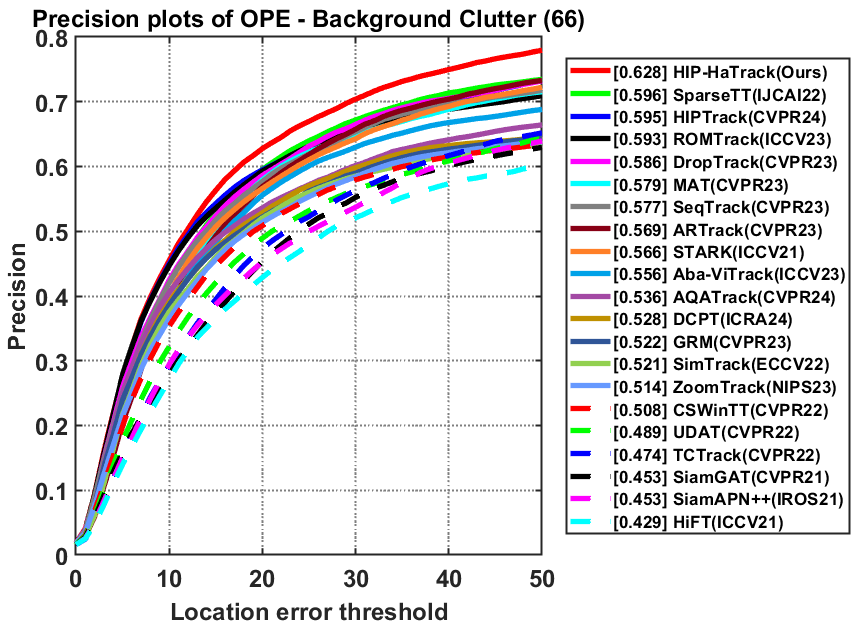}\hspace{0in}
			
	\end{minipage}}
 {
		\begin{minipage}[t]{0.48\textwidth}
			\centering
			\includegraphics[width=1\textwidth,height=0.58\textwidth]{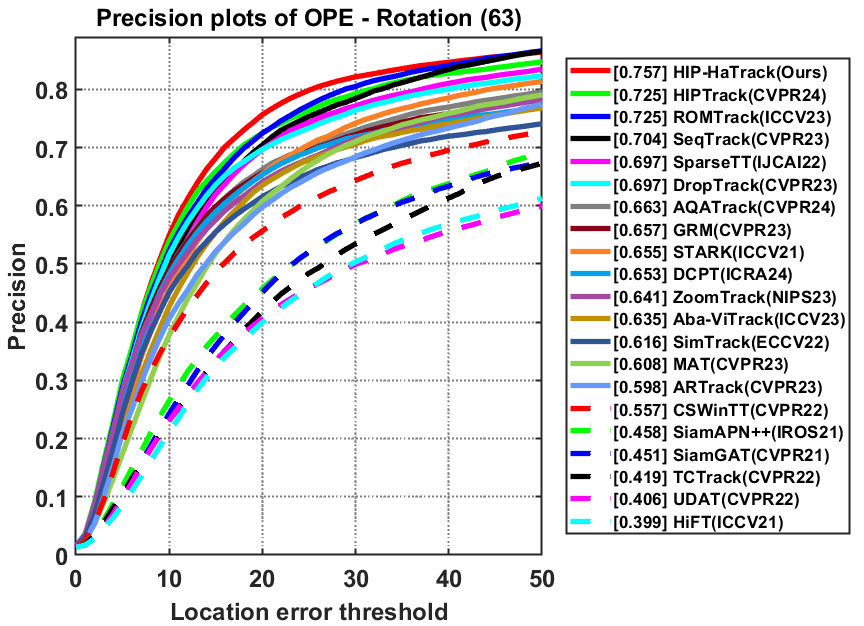}\hspace{0in}
			
	\end{minipage}}
 {
		\begin{minipage}[t]{0.48\textwidth}
			\centering
			\includegraphics[width=1\textwidth,height=0.58\textwidth]{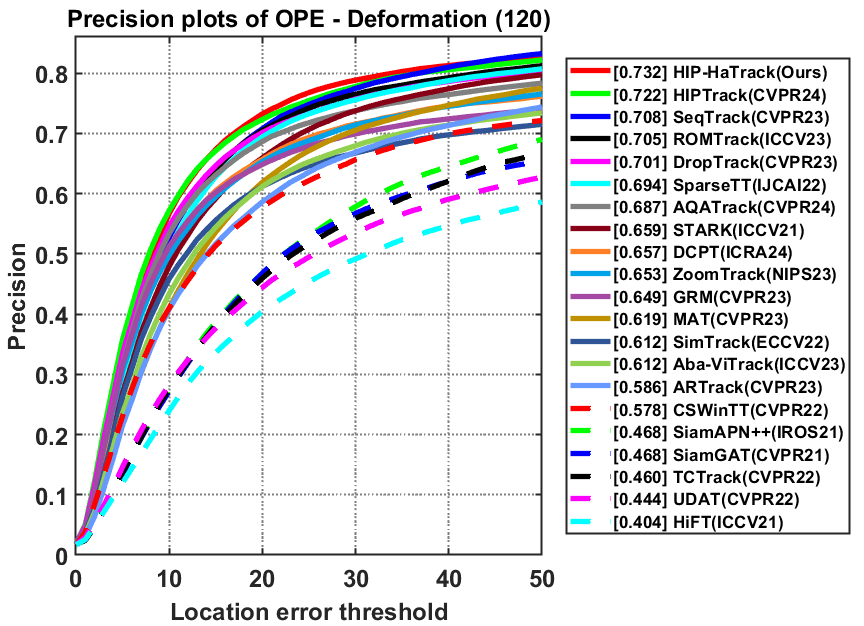}\hspace{0in}
			
	\end{minipage}}
 {
		\begin{minipage}[t]{0.48\textwidth}
			\centering
			\includegraphics[width=1\textwidth,height=0.58\textwidth]{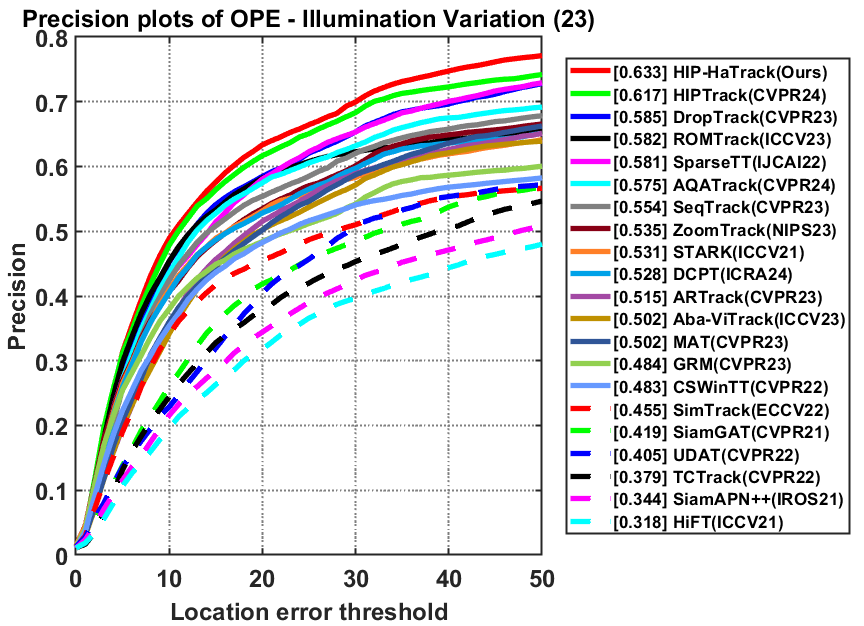}\hspace{0in}
			
	\end{minipage}}
  {
		\begin{minipage}[t]{0.48\textwidth}
			\centering
			\includegraphics[width=1\textwidth,height=0.58\textwidth]{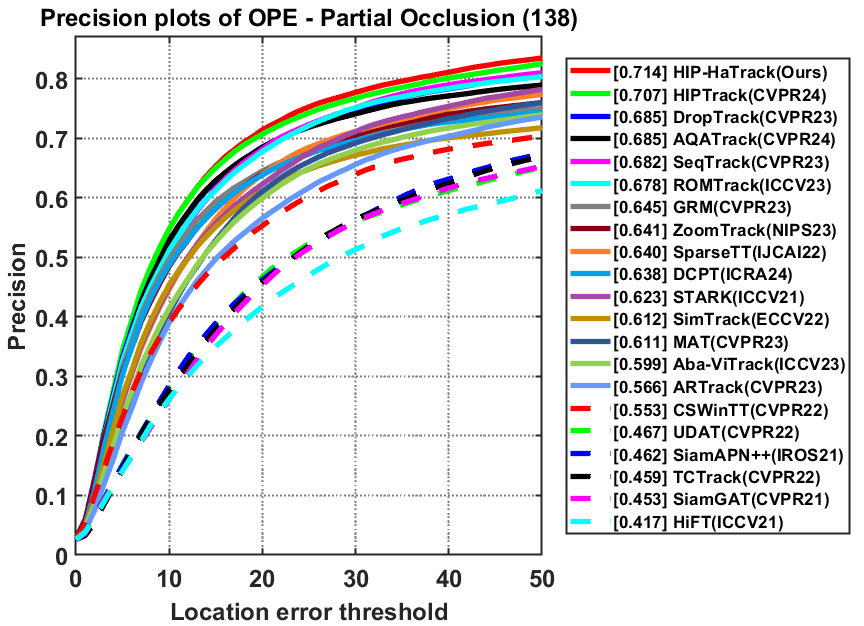}\hspace{0in}
			
	\end{minipage}}
 {
		\begin{minipage}[t]{0.48\textwidth}
			\centering
			\includegraphics[width=1\textwidth,height=0.58\textwidth]{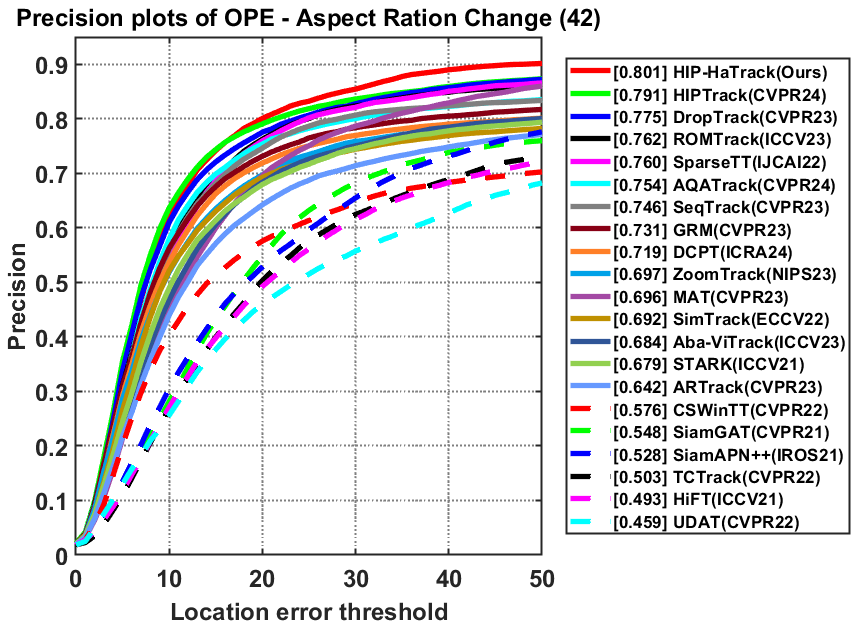}\hspace{0in}
			
	\end{minipage}}
 
 \vspace{-0.15in}
	\caption{Attribute-based comparison on illumination variation, Deformation, Rotation,Background Clutter,Aspect Ration Change and Partial Occlusion. }
	\label{fig:Attribute}\vspace{-0.15in}
\end{figure*}
\noindent\textbf{Success Plot:} Another evaluation metric is the success plot. This metric measures the performance of a tracker based on the overlap of bounding boxes. Given the tracking bounding box $r_{t}$ and the ground truth bounding box $r_{a}$, the overlap score is defined as $S=\frac{\left|r_{t} \bigcap r_{a}\right|}{\left|r_{t} \bigcup r_{a}\right|}$, where $\cap$ and $\cup $ denote the intersection and union of two regions, respectively, and $|\cdot|$ represents the number of pixels in the region. To evaluate the tracking performance over a series of frames, we compute the number of frames with overlap $S$ greater than a given threshold $t_{o}$. The success plot shows the ratio of successful frames at different thresholds between 0 and 1. Using a single success rate value at a specific threshold (e.g. $t_{o}$=0.5) may not be fair or representative for evaluating trackers. Therefore, the area under the curve (AUC) is usually used for each success plot to rank tracking algorithms \cite{53,54}. AUC provides a comprehensive assessment of the overall performance of tracking algorithms across different thresholds, aiding in a more accurate comparison and evaluation of different trackers.

\subsection{Trackers for Comparison}
We evaluate our HIP-HATrack and 20 state-of-the-art trackers to understand their performance on TRO, including HIPTrack \cite{33}, ROMTrack \cite{34}, SeqTrack \cite{35}, DropTrack \cite{36}, AQATrack \cite{37}, SparseTT \cite{38}, DCPT \cite{39}, GRM \cite{52}, Stark \cite{40}, ZoomTrack \cite{41}, MAT \cite{42}, SimTrack \cite{43}, ARTrack \cite{44}, CSWinTT \cite{45}, SiamARN++ \cite{46}, TCTrack \cite{47}, SiamGAT \cite{48}, UDAT \cite{49}, HiFT \cite{50}, and Aba-ViTrack \cite{li2023adaptive}.

\subsection{Evaluation Results}

\noindent\textbf{Overall Performance:}
A comprehensive evaluation of  20  state-of-the-art trackers and our HIP-HATrack was conducted on the TRO dataset. Notably, the existing trackers were used without any modifications for this evaluation. The results are presented with precision and success plots, as displayed in Fig. \ref{fig:TRo_OPE}. As can be seen, HIP-HATrack outperforms all other trackers with a precision (PRC) of 0.738, surpassing its baseline HIPTrack, secuing the second place, with a PRC of 0.723 by 1.5\%. In terms of AUC, HIP-HATrack also leads with the highest AUC of 0.697, surpassing the second place HIPTrack, which has an AUC of 0.689, by 0.8\%. 
These results demonstrate the superior performance of HIP-HATrack in both precision and success rate metrics, highlighting its effectiveness compared to existing state-of-the-art trackers in handling the challenge associated with tracking reflected objects by aggregating hierarchical features.

\noindent\textbf{Attribute-based performance:} We conducted a comprehensive performance evaluation on twelve common attributes to better analyze and understand the performance of different trackers on the TRO dataset. Our HIP-HaTrack achieved the best AUC and PRC for most attributes. Due to the constraints of paper length, we present the PRC results for six common challenges in Fig. \ref{fig:Attribute}, including background clutter (BC), rotation (ROT), deformation (DEF), illumination variation (IV), partial occlusion (POC), and aspect ratio change (ARC).
As can be seen, HIP-HaTrack consistently secures the first place across all attribute subsets, with the baseline HIPTrack taking second place, except for the BC subset where SparseTT ranks second instead. HIP-HaTrack outperforms the second-place trackers by at least 0.7\% on all subsets. Notably, the improvements on the BC and ROT subsets are both 3.2\%. Specifically, HIP-HaTrack achieves a precision of 0.628 on the BC subset and 0.633 on the ROT subset, surpassing SparseTT's 0.596 and HIPTrack's 0.617 by 3.2\%. For the DEF attribute, HIP-HaTrack achieves a precision of 0.732, surpassing HIPTrack's 0.722 by 1.0\%. On the IV, POC, and ARC subsets, HIP-HaTrack shows improvements over HIPTrack by 1.6\%, 0.7\%, and 1.0\%, respectively. These results validate the effectiveness of the proposed method of aggregating hierarchical features for tracking reflective objects. Notably, on the BC and IV subsets, all trackers have precision below 0.650, much lower than the other subsets, which are all above 0.70. This can be attributed to the impact of illumination variation on reflections and the degradation of discriminative cues for reflected objects in cluttered backgrounds, emphasizing the need for specialized techniques to handle reflections.

\begin{figure*}[h]
	\centering
	\includegraphics[width=1\textwidth]{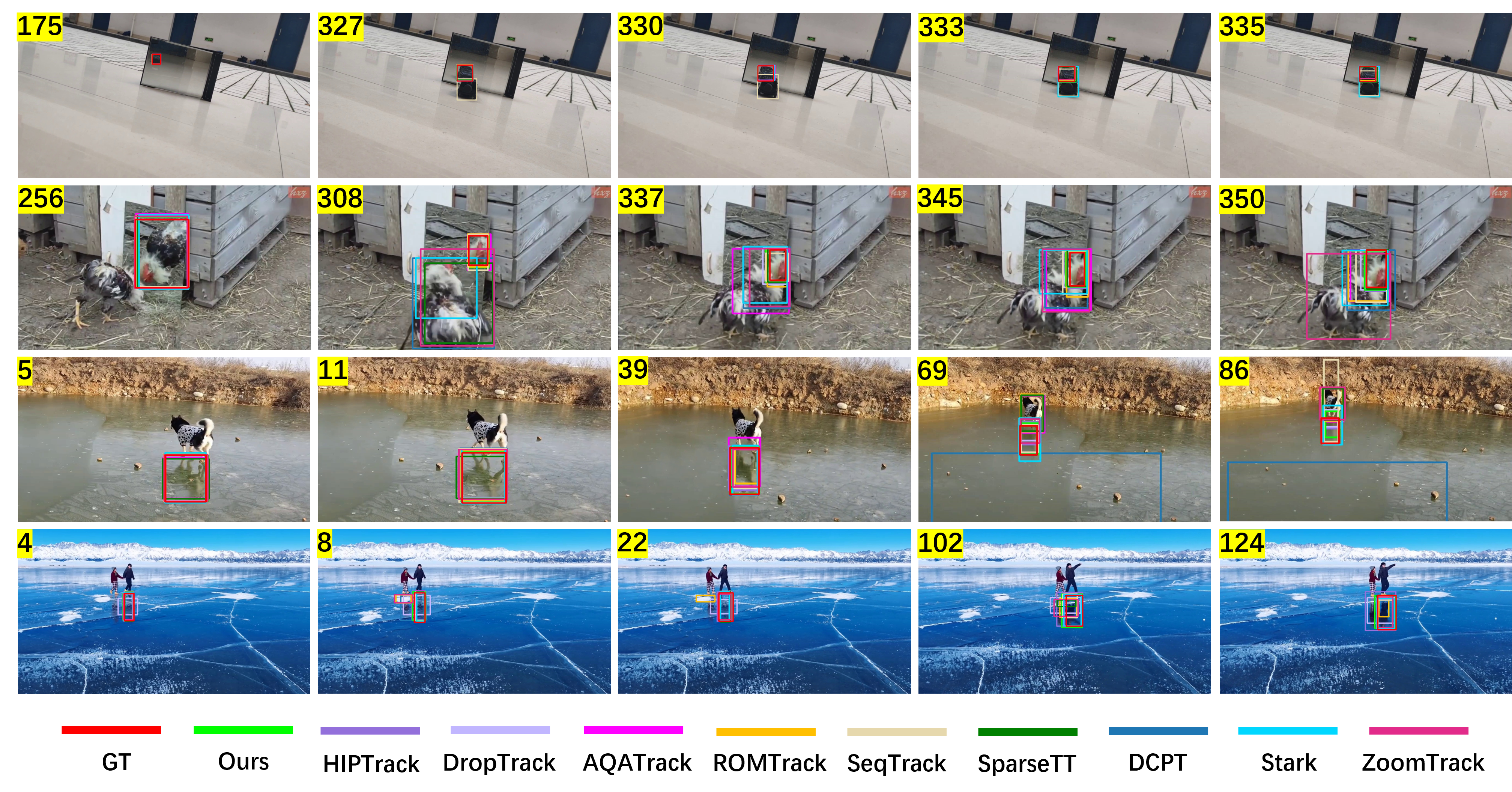}	\vspace{-0.25in}
	\caption{Qualitative evaluation on 4 sequences from TRO, i.e.  Car6,Chicken3,Dog1, and Human112 from top to bottom. The results of different methods have been shown with different colors, and `GT’ denotes the ground truth.}
	\label{fig:Qualitative evaluation}
\end{figure*}

\noindent\textbf{Qualitative evaluation:}  In Fig. \ref{fig:Qualitative evaluation}, we show some qualitative tracking results of our method in comparison with nine top trackers, including HIPTrack \cite{33}, DropTrack \cite{36}, AQATrack \cite{37}, ROMTrack \cite{34}, SeqTrack \cite{35}, SparseTT \cite{38}, DCPT \cite{39}, Stark \cite{40}, and ZoomTrack \cite{41}. These examples are influenced by challenges such as illumination variation, deformation, rotation, background clutter, aspect ratio change, and partial occlusion. 
Specifically, in Car6, only HIP-HaTrack, HIPTrack \cite{33}, DropTrack \cite{36}, ROMTrack \cite{34}, and DCPT \cite{39} successfully track the reflected toy car. In Chicken3, many trackers are influenced by the presence of the real chicken, leading to inaccurate tracking of the reflected one. In Dog1 and Human112, while several trackers manage to track parts of the reflected object, HIP-HaTrack consistently shows visually superior results. These qualitative comparisons support the effectiveness and superiority of our method of aggregating hierarchical features for tracking reflected objects.

\begin{table*}[t]
\centering
\caption{Illustrates the impact of the importance coefficient of aggregated feature on HIP-HaTrack's PRC (\%) and AUC (\%) on TRO. Red, blue, and green represent the first, second, and third place, respectively.}
\label{tab:4}
\vspace{-0.1in}
\resizebox{3in}{0.3in}{%
\begin{tabular}{@{}cccccccccccc@{}}
\toprule
$\rho$ & 1.0 & 0.9 & 0.8 & 0.7 & 0.6 & 0.5 & 0.4 & 0.3 & 0.2 & 0.1 \\ \midrule
PRC (\%) & 68.6 & 68.6 & 69.1 & 70.6 & {\color{blue} \textbf{72.3}} & 71.0 & {\color{green} \textbf{71.8}} & {\color{red} \textbf{73.8}} & 71.2 & 71.1 \\
AUC (\%) & 67.9 & 68.0 & 68.4 & {\color{blue} \textbf{69.3}} & {\color{green} \textbf{69.0}} & 68.1 & 67.9 & {\color{red} \textbf{69.7}} & 69.0 & 68.9 \\ \bottomrule
\end{tabular}%
}
\vspace{-0.12in}
\end{table*}
\subsection{Ablation Study}
\noindent\textbf{Impact of the importance coefficient of aggregated hierarchical feature}. To study the impact of the weight coefficient $\rho$ on balancing the contributions of aggregated hierarchical feature and $H^{\downarrow }_n$ in HIP-HaTrack, we evaluated HIP-HaTrack on the TRO dataset with various $\rho$ settings. Specifically, $\rho$ ranges from 0.1 to 1.0 with step of size 0.1. It is important to note that as $\rho$ increases, the contribution of the aggregated feature increases, while the contribution of $H^{\downarrow }_n$ decreases. Table \ref{tab:4} shows HIP-HaTrack's PRC and AUC on the TRO dataset under different $\rho$ values. The results indicate that PRC and AUC reach their highest values when $\rho$ is between 0.3 and 0.7, highlighting the significance of aggregated hierarchical features for improving tracking performance. Specifically, when $\rho$ = 0.3, HIP-HaTrack achieved the best PRC and AUC, with values of 0.738 and 0.697, respectively. Therefore, we used $\rho$=0.3 as the default setting for HIP-HaTrack.

\section{Conclusion}

In this work, we investigate a relatively unexplored tracking task, namely, tracking reflected objects. Reflections can significantly distort the appearance of objects, creating ambiguous visual cues that complicate the tracking process. This issue is particularly pertinent in applications such as autonomous driving, security, smart homes, and industrial production, where accurately tracking objects reflected in surfaces like mirrors or glass is crucial. However, there hasn't been public benchmark dedicated to tracking reflected objects. 
To address this gap, we introduce TRO, which, to the best of our knowledge, is the first benchmark specifically designed for reflected object tracking. Additionally, to assess the performance of existing trackers and establish a baseline for future comparisons, we extensively evaluate 20 state-of-the-art tracking algorithms with in-depth analysis. Furthermore, we propose a novel tracker named HiP-HaTrack, which aggregates hierarchical features to achieve better feature representations, significantly outperforming existing state-of-the-art algorithms.

\clearpage 

\bibliographystyle{splncs04}
\bibliography{main}

\end{document}